\documentclass[journal]{IEEEtran}
\usepackage{cite}

\ifCLASSINFOpdf
   \usepackage[pdftex]{graphicx}
\else
\fi
\usepackage{amsmath}
\usepackage{amsfonts}
\usepackage{setspace}

\usepackage{soul} 
\usepackage{booktabs} 

\usepackage[colorlinks=true,linkcolor=blue,citecolor=blue,urlcolor=black ]{hyperref}

\usepackage{graphicx}

\usepackage{subcaption}

\begin{document}
%
\title{Robust Ultra-wideband Range Error Mitigation with Deep Learning at the Edge}
%
%
%

\author{Simone~Angarano, Vittorio~Mazzia, Francesco~Salvetti,
        Giovanni~Fantin and~Marcello~Chiaberge
\thanks{The authors are with Politecnico di Torino -- Department of Electronics and Telecommunications, PIC4SeR, Politecnico di Torino Interdepartmental Centre for Service Robotics and SmartData@PoliTo, Big Data and Data Science Laboratory, Italy. Email: \{name.surname\}@polito.it.}}

%
%

\markboth{}%
{Simone Angarano \MakeLowercase{\textit{et al.}}: Robust Ultra-wideband Range Error Mitigation with Deep Learning at the Edge}
%



\maketitle

\begin{abstract}
Ultra-wideband (UWB) is the state-of-the-art and most popular technology for wireless localization. Nevertheless, precise ranging and localization in non-line-of-sight (NLoS) conditions is still an open research topic. Indeed, multipath effects, reflections, refractions, and complexity of the indoor radio environment can easily introduce a positive bias in the ranging measurement, resulting in highly inaccurate and unsatisfactory position estimation. This article proposes an efficient representation learning methodology that exploits the latest advancement in deep learning and graph optimization techniques to achieve effective ranging error mitigation at the edge. Channel Impulse Response (CIR) signals are directly exploited to extract high semantic features to estimate corrections in either NLoS or LoS conditions. Extensive experimentation with different settings and configurations has proved the effectiveness of our methodology and demonstrated the feasibility of a robust and low computational power UWB range error mitigation.
\end{abstract}


%

\section{Introduction}  
Precise localization is at the core of several engineering systems, and due to its intrinsic scientific relevance, it has been extensively researched in recent years \cite{zafari2019survey, vo2015survey}. Either outdoor or indoor applications could largely benefit from it in fields as diverse as telecommunications \cite{menta2019performance}, service robotics \cite{karlsson2004core}, healthcare \cite{cheng2016multiple}, search and rescue \cite{zorn2010novel} and autonomous driving \cite{jo2013gps}. Nevertheless, accurate positioning in non-line-of-sight (NLoS) conditions is still an open research problem. Multipath effects, reflections, refractions, and other propagation phenomena could easily lead to error in the position estimation \cite{stahlke2020nlos,wen2019gnss,ray2013subseasonal}.

\begin{figure}[b]
    \centering
    \includegraphics[scale=0.55]{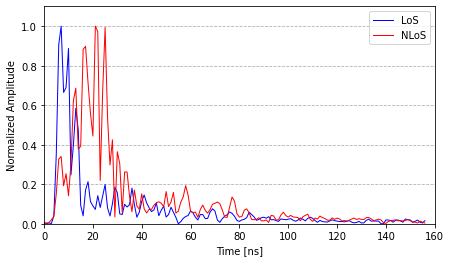}
    \caption{LoS and NLoS CIRs with normalized amplitude in an indoor environment. In the NLoS case, the signal travels along many routes until it reaches the antenna. That makes the ToA estimation ambiguous.}
    \label{fig:nlos_vs_los}
\end{figure}

Ultra-wideband (UWB) is the state-of-the-art technology for wireless localization, rapidly growing in popularity \cite{tiwari2020design}, offering decimeter-level accuracy and increasingly smaller and cheaper transceivers \cite{magnago2019robot}. With a bandwidth larger than 500 MHz and extremely short transmit pulses, UWB offers high
temporal and spatial resolution and considerable multipath effect error mitigation when compared to other radio-frequency technologies \cite{schmid2019accuracy}. Nevertheless, UWB is still primarily affected by the NLoS condition, Fig. \ref{fig:nlos_vs_los}, in which the range estimates based on time-of-arrival (TOA) is typically positively biased \cite{savic2015kernel, otim2019effects}. That is particularly true for indoor localization, where ranging errors introduced by multipath and NLoS conditions can quickly achieve large deviations from the actual position \cite{chen2020uwb}.
So, robust and effective mitigation is necessary to prevent large localization errors.

Several approaches have been proposed to address the NLoS problem. In the presence of a large number of anchor nodes available, NLoS identification is the preferable choice so far. Indeed, once an NLoS anchor is identified, it can be easily eliminated from the pool of nodes used for the trilateration algorithm \cite{silva2016ir}. The majority of the proposed methodologies found in the literature make use of channel and waveform statistics \cite{marano2010nlos,schroeder2007nlos,barral2019nlos}, likelihood ratio tests or binary hypothesis tests \cite{silva2016ir,muqaibel2013practical} and machine learning techniques. In the latter case either hand-designed techniques, such as support vector machine (SVM), \cite{ying2012classification}, Gaussian processes (GP), \cite{xiao2014non}, or representation learning models have been investigated \cite{jiang2020uwb,stahlke2020nlos}.

\begin{figure*}
    \centering
    \includegraphics[height=3.8cm]{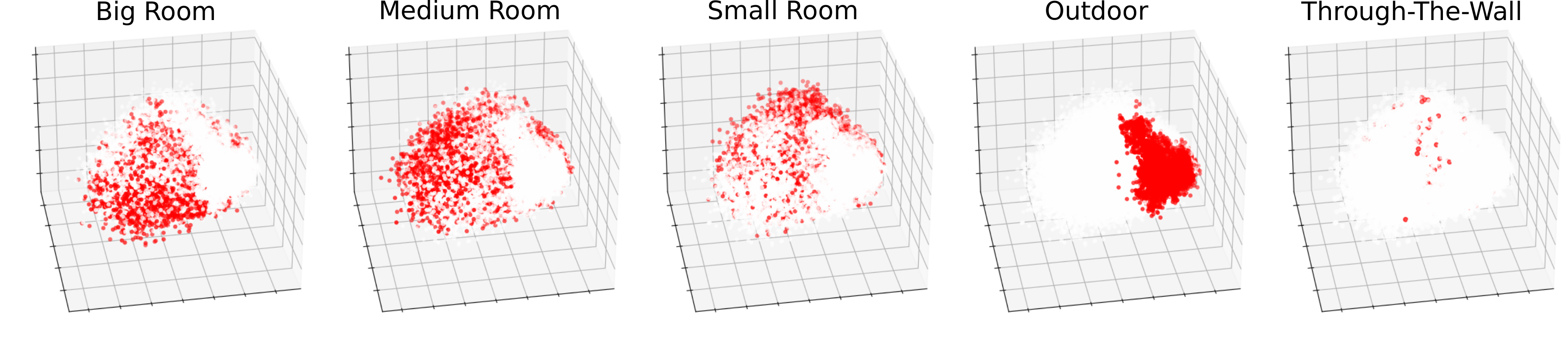}
    \includegraphics[height=3.8cm]{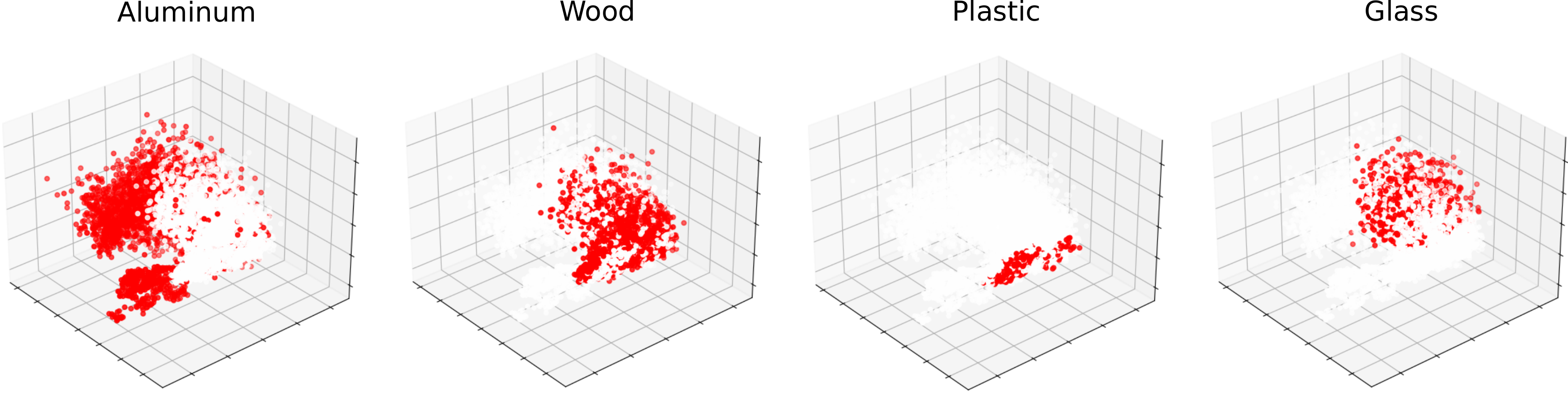}
    \caption{Principal Component Analysis (PCA), projecting the original 157 CIR dimensions into a three-dimensional space. It is clear how rooms cover a similar data-space, completely separated by the outdoor scenario. Moreover, the same apply to materials, where more dense molecular structures affect the signal differently.}
    \label{fig:PCA_rooms}
\end{figure*}
Despite the simplicity of applying NLoS identification, \cite{gururaj2017real}, in almost all practical situations, there is no sufficient number of anchors available to exclude some of them. So, the majority of research community efforts focus on range mitigation and direct localization mitigation. Regarding the latter, even if there are studies that show excellent position estimation in multipath environments, \cite{niitsoo2018convolutional,hsieh2019deep,poulose2020uwb}, the collected training data are incredibly site-specific. Therefore, conducting the data collection on one site does not allow exploiting the resulting model in another location.
On the other hand, range mitigation is far less site-specific and does not require a large amount of data to achieve satisfactory results \cite{savic2015kernel}. Range error mitigation is mostly performed with similar techniques as NLoS identification \cite{xiao2014non, zeng2019nlos, mao2018probabilistic,wymeersch2012machine} and also with more extreme conditions as error mitigation for through-the-wall (TTW) \cite{silva2020ranging}. Moreover, following the advancements bring by representation learning techniques in many fields of research \cite{khaliq2019refining,aghi2020local,salvetti2020multi}, Bregar \& Mohorčič attempted to perform range error estimation directly from the channel impulse response (CIR) using a deep learning model \cite{bregar2018improving}. Nevertheless, being a preliminary study, no relevance has been given to studying the network, optimizing it, and making it able to generalize to different environments.

This article focuses specifically on investigating a novel efficient deep learning model that performs an effective range error mitigation, using only the raw CIR signal at the edge. Indeed, range error mitigation should be performed directly on the platform where the UWB tag is attached.  So, energy consumption and computational power play a decisive role in the significant applicability of our methodology. We adopt the latest advancements in deep learning architectural techniques \cite{wang2017deep,szegedy2016inception}, and graph optimization \cite{jacob2018quantization} to improve nearly 45\% and 34\% the NLoS and LoS conditions, respectively, in an unknown indoor environment up to barely 1 mJ of energy absorbed by the network during inference. Moreover, our proposed methodology does not require additional NLoS identification models. Still, it is able to extract valuable features to estimate the correct range error directly from the CIR in both LoS and NLoS states. The main contributions of this article are the following.

\begin{enumerate}
    \item Design and train a highly efficient deep neural architecture for UWB range mitigation in NLoS and LoS conditions using only raw CIR data points.
    \item Introduce weight quantization and graph optimization for power and latency reduction in range error mitigation.
    \item Evaluate and compare several devices and hardware accelerators, annotating power and computational request for different optimized networks.
    \item Collect and analyze a novel open-source dataset with different NLoS scenarios and settings, highlighting the generalization limitations of a hierarchical learning model.
\end{enumerate}

All of our training and testing code and data are open source and publicly available\footnote{\hl{https://zenodo.org/record/4399187}}.

The rest of the paper is organized as follows. Section \ref{dataset_construction} covers the dataset creation and the preliminary analysis conducted on generic learning algorithms for ranging error mitigation to assess their ability to provide a generalized data representation.
Section \ref{methodology} presents a detailed explanation of the efficient REMnet deep learning architecture and the proposed graph and quantization techniques used to achieve a significant range correction for the trilateration algorithm in LoS and NLoS conditions. Finally, Section \ref{experiments_results} presents the experimental results and discussion, followed by the conclusion.

\begin{figure*}
    \centering
    \includegraphics[width=0.9\linewidth]{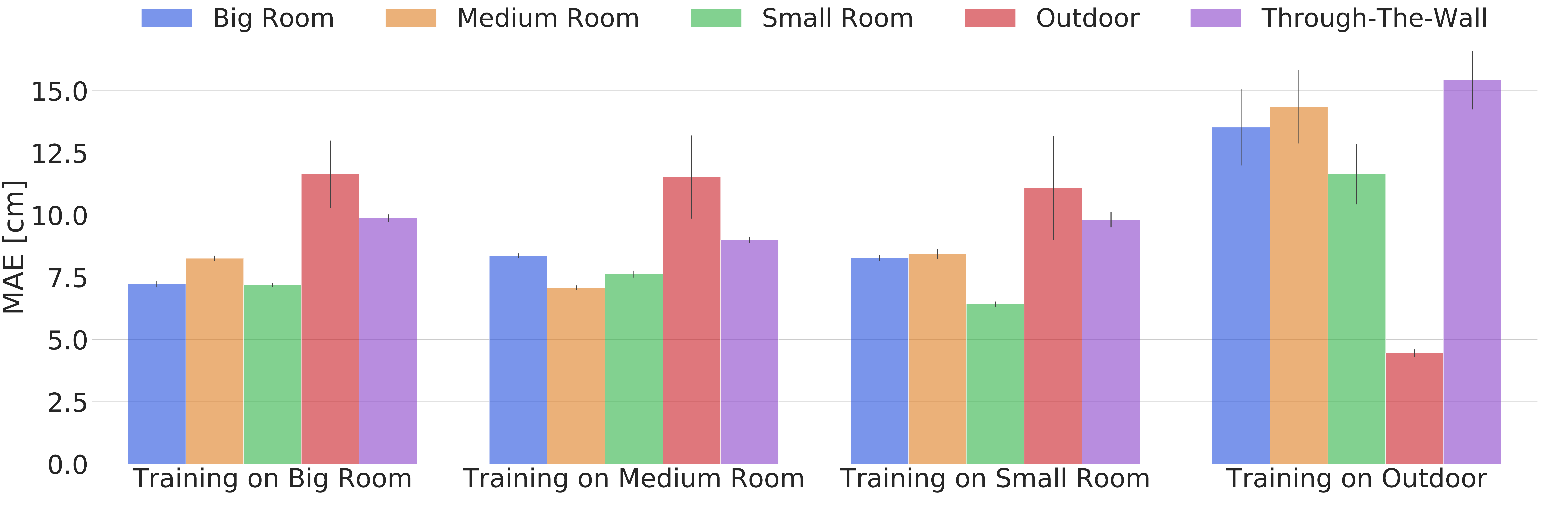}
    \caption{Analysis of generalization capabilities in different environments of a generic representation learning model trained directly on the CIR waveform. Each bar represents the average of 20 independent MLP models. Results show that rooms with different sizes and configurations lead to very similar losses. Moreover, outdoor and TTW scenarios should be considered as separate settings and cannot be corrected without including appropriate samples in the training set.}
    \label{fig:bars}
\end{figure*}

\section{Dataset Construction}
\label{dataset_construction}
The measurements are taken in five different environments to cover a wide variety of LoS and NLoS scenarios: an outdoor space, in which the only source of error is the presence of obstacles, and three office-like rooms, to include the effect of multipath components. In particular, the biggest room is approximately 10m x 5m large, the medium one is 5m x 5m, and the smallest is 5m x 3.5m. Moreover, to analyze the TTW effect, some measurements are acquired across different rooms. Taking range measurements in different conditions allows performing training, validation, and testing on entirely different datasets, avoiding overfitting and encouraging a representation learning model to learn domain-independent features.

The EVB1000 boards are configured to guarantee precise ranging and high update frequency according to the constructor's manual, and antenna delays are tuned to compensate for measurement bias. The measurements are taken using a Leica AT403 laser tracker as ground truth. First, we measure the anchor's position to have a landmark; then, the laser follows the reflector placed on the moving tag estimating its position ten times per second. Meanwhile, tag and anchor perform two-way ranging at approximately the same frequency. The tag follows a path in an environment filled with obstacles to generate both LoS and NLoS measurements. After a satisfying number of samples are obtained, the configuration is changed by modifying the anchor's position or the type and position of the obstacles.

To compute the ranging error, we match each measure with the ground-truth range from the laser tracker comparing timestamps. Each of the 55,000 samples of the dataset also contains the CIR vector, giving information about the transmission channel of the UWB signal. For each vector, only 152 samples after the first detected peak are retained, as suggested in \cite{Bregar2018}. Moreover, five additional samples before the peak are included to compensate for eventual errors in the detection. Finally, the environment and obstacles used for the measurements are reported to study their effect on the proposed method. As previously stated, the whole dataset is publicly released to be useful for future works on this subject \cite{DeepUWB_Dataset}.

\begin{figure*}
    \centering
    \includegraphics[width=1\linewidth]{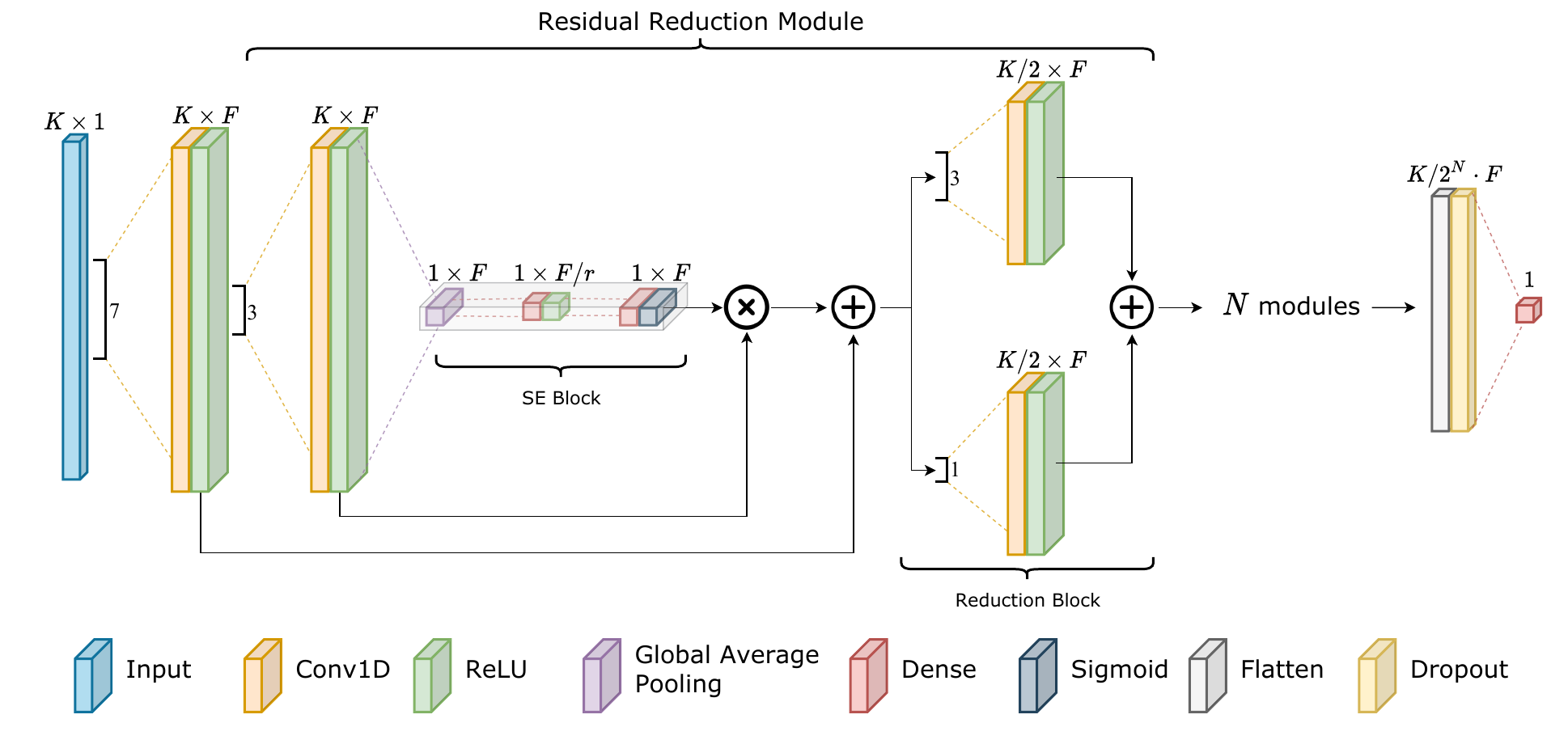}
    \caption{Overview of the REMnet architecture. The input of the model is the $K\times1$ tensor representing the CIR of the measurement. The dimensionality is reduced by $N$ subsequent Residual Reduction Modules (RRM) with a feature attention mechanism. Finally, a fully connected layer composes the high-level extracted features and outputs the range error estimation.}
    \label{fig:architecture}
\end{figure*}

\subsection{Dataset analysis}
\label{dataset_analysis}

To visualize the distribution of the acquired instances in the data space, we exploit Principal Component Analysis (PCA) to project the 157 dimensions of each CIR signal into a three-dimensional space, saving most of the original variance. As shown in Fig. \ref{fig:PCA_rooms}, the first analysis highlights the correlations between data points in the different analyzed environments. A prevalence of samples from the big room can be found in the lower central part of the plot, while the medium and small room samples are more present in the left and upper side of the distribution, respectively. Nevertheless, it is clear how rooms cover a similar data space, which implies a potential transferability of statistics learned in different indoor environments. On the other hand, the outdoor set is completely separated and wholly concentrated on the right side of the plot.

The same procedure is followed for materials, considering four object classes for clearness: aluminum plate, plastic bins, wooden door, and glass. In this case, a remarkable separation is noticeable, as the metal samples occupy all the left part of the graph and light objects like plastic, wood, and glass take the right area. Moreover, the spatial distribution of wood occupies specific zones showing different features from plastic and glass.
The presented qualitative analysis allows a first visual proof of the meaningfulness of data and draws some conclusions on how a representation learning model could perform. For example, a generic model trained on measures taken with only plastic instances would more easily mitigate the error caused by wood and less accurate estimations for metal samples.

Finally, a Multilayer Perceptron (MLP) is trained and tested on different splits of the dataset to assess the generalization capabilities of a generic representation learning model trained directly on the CIR waveform.
After the validity of the method is first verified on the whole dataset, a series of tests are conducted to study the effect of different environments and obstacles on the models' performance. The network is trained on a specific set of data from the same setting or material and tested on other possible scenarios. In this way, it is possible to state whether the approach holds an absolute generality concerning such factors. 
For what concerns environmental influence, resumed in Fig. \ref{fig:bars}, metrics show that rooms with different sizes and configurations lead to minimal losses (less than 2 cm) compared to those caused by outdoor measurements or more extreme conditions as TTW. Indeed, samples taken in open space show the worst results because they are taken in a completely different scenario. So, models struggle to adapt to a situation in which multipath components are absent, but an improvement is achieved in almost all cases.
Regarding obstacles, we can notice a more marked distinction. As already emerged from PCA analysis, heavy materials have a very different impact on UWB signals with respect to wood, plastic, and glass. However, there is almost always an improvement in the raw mean absolute error (MAE). That means that models can learn a way to compensate part of the error independently from the obstacles. A dataset containing a sufficient number of examples for a wide variety of materials can lead to excellent results in many different scenarios.

\section{Proposed Methodology}
\label{methodology}

In this section, we propose a Deep Neural Network (DNN) to solve the range error mitigation problem. Moreover, we present some optimization and quantization techniques used to increase the computational efficiency of the network. Since UWB are low-power localization devices directly connected to the mobile robot board, any error compensation technique should be applied locally on the platform to ensure real-time execution with a latency compatible with the control frequency of the robot. The method should also be as efficient as possible to ensure a low impact on the system's overall energy and computational demand. In designing our solution, we mainly focus on optimizing the model to reduce memory occupancy and computational efforts during inference.

\subsection{Network Design}
We consider the following model for a generic UWB range measurement:
\begin{equation*}
    \hat{d} = d +  \Delta d
\end{equation*}
 \noindent where the actual distance $d$ is intrinsically affected by an error $\Delta d$ giving the final measurement outcome $\hat{d}$. The error depends on several factors, among which the most important is the environment and the obstacles, giving, in general, worse performance in NLoS condition.
 
 We formulate the mitigation problem as a regression of the compensation factor $\Delta d$ that should be added to the measured range to obtain the actual distance between the two sensors. Therefore, we design a DNN model that predicts an estimate $\hat{y}$ for the true latent error $y=\Delta d$ as a non-linear function of the input CIR vector $X$. We call the proposed architecture Range Error Mitigation network (REMNet). It is essential to underline that we do not distinguish between LoS and NLoS measurements, but we let the network learn how to compensate for both the conditions autonomously. Therefore, a classification of the measurements is not computed, but the model implicitly performs it during the mitigation process. Such an approach allows to obtain an algorithm that is always beneficial and can be applied continuously on-board without the need for an additional classification step.

Due to the one-dimensional nature of the data, we select 1D convolutional layers as building blocks of the network.We denote with $K$ the number of temporal samples of the input CIR vector $X$. We first extract $F$ low-level features with a 1D convolution. The network architecture is then made of a stack of $N$ Residual Reduction Modules (RRM) that learn deep features from the high-level features while reducing the temporal dimensionality $K$. We develop this module adopting well-known strategies used in deep learning literature such as residual connections \cite{he2016deep}, attention mechanism \cite{vaswani2017attention, hu2018squeeze, woo2018cbam} and sparsely connected graphs \cite{szegedy2015going}. All these methodologies have been proven to be effective to guarantee trainable and well converging networks and are therefore suitable to be applied with the range error mitigation problem.

The core of the RRM is composed of a residual unit followed by a reduction block:

 \begin{equation*}
    \text{RRM}(x)  = \text{Red}(\text{Res}(x))
\end{equation*}

The residual unit has a 1D convolution followed by a Squeeze-and-Excitation (SE) block \cite{hu2018squeeze}on the residual branch:

 \begin{equation*}
    \text{Res}(x) = \text{SE}(\text{Conv1D}(x)) + x
\end{equation*}

The SE block applies a feature attention mechanism by self-gating each extracted feature with a scaling factor obtained as a non-linear function of themselves. Denoting with $x$ the $K\times F$ tensor of feature maps extracted by the convolutional layer, we first squeeze it with a global average pooling layer that aggregates the tensor along the temporal dimension, obtaining a single statistic for each feature. The excitation step is then performed with a stack of one bottleneck fully connected (FC) layer that reduces the feature dimension $F$ of a factor $r$ and another FC layer that restores the dimensionality to $F$ with sigmoid activation. This activation outputs $F$ independent scaling factors between 0 and 1 that are then multiplied with the input $x$, allowing the network to focus on the most prominent features. Overall, the SE output is computed as:

\begin{equation*}
    \text{SE}(x) = \text{FC}_2\Big(\text{FC}_1\Big(\frac{1}{K} \sum_{i} x_{ij}\Big)\Big) \cdot x
\end{equation*}

where

\begin{equation*}
\begin{split}
    \text{FC}_1(x) &= \text{max}(0,xW_1+b_1) \quad,\quad W_1\in\mathbb{R}^{F\times F/r}\;,\;b_1\in\mathbb{R}^{F/r}\\
    \text{FC}_2(x) &= \text{sigmoid}(xW_2+b_2) \quad,\quad W_2\in\mathbb{R}^{F/r\times F}\;,\;b_1\in\mathbb{R}^{F}
\end{split}
\end{equation*}

The residual unit is followed by a reduction block, which halves the temporal dimension $K$ with two parallel convolutional branches with a stride of 2:

 \begin{equation*}
    \text{Red}(x) = \text{Conv1D}_1(x) + \text{Conv1D}_2(x)
\end{equation*}

where both $\text{Conv1D}_1$ and $\text{Conv1D}_2$ have $F$ channels, but different kernel size in order to extract different features.

After $N$ Residual Reduction Modules, we end up with a tensor with shape $K/2^N\times F$. We flatten it into a single vector, and we apply a Dropout layer to avoid overfitting and help generalization. Finally, an FC layer with linear activation computes an estimate of the compensation value $\Delta d$. Except for this final layer and the second FC layer in the SE blocks, we always apply a ReLU non-linearity to all the layers. All the convolutional layers are also zero-padded so that the temporal dimension is reduced by the stridden convolutions of the reduction block, only. An overview of the overall network architecture is presented in Fig. \ref{fig:architecture}.

\subsection{Network optimization and quantization techniques}
As already mentioned, a UWB range error mitigation technique should respect constraints on memory, power, and latency requirements to be applicable in real-time and on-board. For this reason, we investigate different graph optimization and quantization methods to both decrease model size and computational cost. In the literature, several techniques to increase neural network efficiency can be found \cite{gong2014compressing,gupta2015deep,han2016eie,han2015deep,jacob2018quantization}. In particular, we focus on the following main approaches:
\begin{itemize}
    \item network pruning and layer fusing that consists in optimizing the graph by removing low-weight nodes that give almost no contribution to the outcome and fuse different operations to increase efficiency;
    \item weights quantization that consists of reducing the number of bits required to represent each network parameters;
    \item activations quantization, that reduces the representation dimension of values during the feed-forward pass, thus reducing also the computational demand;
    \item quantization-aware training, in which the network is trained considering a-priori the effect of quantization trying to compensate it.
\end{itemize}

We produce five different versions of REMnet, depending on the adopted techniques. The first is the plain float32 network with no modifications. We apply graph optimization to this first model without quantization to investigate its effect on precision and inference efficiency. The third version is obtained by quantizing the weights to 16 bits, while activations and operations are still represented as 32 bits floating points. The last two models deal instead with 8 bits full integer quantization.

This strategy is the most radical to increase network efficiency by changing the representation of both weights and activations to integers, greatly reducing memory and computational demands due to the high efficiency of integer computations. However, a great problem is how to manage completely by integer-only operations the feed-forward pass of the network. We follow the methodology presented by Jacob et al. \cite{jacob2018quantization} in which each weight and activation are quantized with the following scheme:

\begin{equation*}
    r = S(q-Z)
\end{equation*}

\noindent where $r$ is the original floating-point value, $q$ the integer quantized value, and $S$ and $Q$ are the quantization parameters, respectively scale and zero point. A fixed-point multiplication approach is adopted to cope with the non-integer scale of $S$. Thus, all computations are performed with integer-only arithmetic making inference possible on devices that do not support floating-point operations. We obtain two full-integer models by adopting both post-training quantization and quantization-aware training. With this second approach, fake nodes are added to the network graph to simulate quantization effects during training. In this way, the gradient descent procedure can consider the integer loss in precision.

All the inference results obtained with the five models are presented in Section \ref{quantitative_results}.  

\section{Experiments and Results}
\label{experiments_results}

In this section, we perform an experimental evaluation of the proposed neural efficient architecture for range error mitigation. Moreover, we test the accuracy and performance of different optimized versions of the network on disparate heterogeneous devices collecting energy and computational requirements.

\begin{figure}
    \centering
    \includegraphics[scale=0.55]{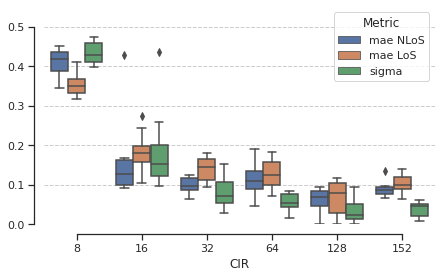}
    \caption{Network performances with different CIR sizes $K$, starting from the dimension suggested by \cite{bregar2018improving}. Progressively training with a reduced number of input features degrades the performance of the network. An input with eight dimensions appears to be the minimum required to obtain an acceptable range error estimation.}
    \label{fig:cir_reduction}
\end{figure}

\subsection{Experimental setting}
\label{experimental_setting}

In the following experiments, we employ the presented dataset of Section \ref{dataset_construction} keeping aside the medium size room as the testing set. Indeed, instead of performing a stratified sampling of the available data, in the light of the evidence of Section \ref{dataset_analysis}, we decide to perform all tests with indoor instances. That is more similar to an actual infield application and better highlights the generalization capabilities of the proposed methodology. All experimentations are performed with 36023 and 13210 training and testing data points, respectively, keeping aside TTW and outdoor measurements. Finally, due to their very different nature, and explicitly labeled LoS samples are employed to evaluate the network's capability to recognize this condition and act accordingly.

The final test consists in using the best-developed model for a 3D positioning task to assess range mitigation's effect on localization accuracy. The medium room is chosen as the testing environment, as its samples have not been used to train the networks. Four UWB anchors are placed in the room, and a fixed tag is put in the center. First, the laser tracker precisely measures the position of all the nodes to provide ground truth, then the acquisition of the data begins. Two situations are taken into consideration, a fully LoS scenario and a critical NLoS one. Once the samples have been collected, they are prepared for the processing phase, in which range measurements are used to estimate the 3D position of the tag employing a simple Gauss-Newton non-linear optimization algorithm.

All network hyperparameters are obtained with an initial random search analysis followed by a grid search exploration to fine-tune them and find a compromise between accuracy and efficiency. Indeed, working at the architecture level is crucial to satisfy the constraints given by the studied application. The number of filters, $F$, is equal to 16 and the number of reduction modules $N=3$ with $r=8$. As shown in Fig. \ref{fig:architecture}, all 1D convolutional operations have a kernel of size 3, except for the first layer and the second branch of the reduction block. The resulting network has an efficient and highly optimized architecture with 6151 trainable parameters. Finally, to select the optimal number of input features, as shown in Fig. \ref{fig:cir_reduction}, we progressively reduced the input number of dimensions $K$ while annotating the network metrics. All points are the average result of ten consecutive trials. Experimentation shows that eight dimensions are the minimum number of features required to the network to obtain an acceptable range error estimation. Moreover, we empirically find that an input CIR of 152 elements, as suggested by \cite{bregar2018improving}, is redundant and could even slightly reduce the model's performance. On the other hand, fewer dimensions of 128 tend almost linearly to degrade the network's accuracy.

The Adam optimization algorithm \cite{kingma2014adam} is employed for training, with momentum parameters $\beta_{1}=0.9$, $\beta_{2}=0.999$, and $\epsilon=10^{-8}$. The optimal learning rate, $\eta=3e-4$, is experimentally derived using the methodology described in \cite{smith2017cyclical}. That is kept constant for 30 epochs with a batch size of 32 and MAE loss function. We employ the TensorFlow framework to train the network on a PC with 32-GB RAM and an Nvidia 2080 Super GP-GPU. The overall training process can be performed in less than 10 minutes.

\begin{table*}[]
\centering
\begin{tabular}{llllll}
\toprule
                                   & MAE {[}NLoS{]} & MAE {[}LoS{]} & $R^2$ {[}NLoS{]} & $R^2$ {[}LoS{]} & $\sigma$ {[}NLoS{]} \\
                                   \hline
Support Vector Machine (SVM)             & 0,0766         & 0,0507        & 0,4444                          & 0,1256                         & 0,1171         \\ 
Multilayer Perceptron (MLP)             & 0.0796         & 0.0466        & 0.4194                          & 0.2913                         & 0.1170         \\ 
CNN-1D \cite{bregar2018improving}            & 0.0890         & 0.0523        & 0.2870                          & 0.1089                         & 0.1285         \\ \hline \hline
REMnet float32                  & 0.0687         & 0.0445        & 0.5607                          & 0.3483                         & 0.1057         \\ 
Graph Optimization                 & 0.0687         & 0.0445        & 0.5607                          & 0.3483                         & 0.1057         \\ 
Post-training float16 quantization & 0.0688         & 0.0445        & 0.5607                          & 0.3484                         & 0.1058         \\ 
Post-training 8-bit quantization   & 0.0712         & 0.0455        & 0.5361                          & 0.3100                         & 0.1082         \\ 
Full-integer aware quantization    & 0.0708         & 0.0449        & 0.5404                          & 0.3357                         & 0.1079         \\ \bottomrule
\end{tabular}
\caption{Proposed architecture performances after graph optimization and different levels of weight quantization. Initial values of MAE for NLoS and LoS sginals are 0.1242 m and 0.0594, respectively. It is possible to notice how the different transformations barely affect the range error estimation capability of the network. Moreover, three baseline approaches are tested and compared with the efficient REMnet model and its optimization versions.}
\label{tab:optimization_results}
\end{table*}

\subsection{Quantitative results}
\label{quantitative_results}

\begin{figure}
    \centering
    \includegraphics[scale=0.55]{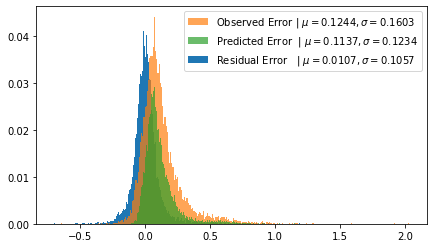}
    \caption{Normalized histograms with 300 bins each. It is possible to notice how the residual range error distribution (blue histogram) is almost Gaussian. That greatly improves the optimality and simplicity of the subsequent iterative localization algorithm \cite{sarkka2013bayesian}.}
    \label{fig:hist_pred_res}
\end{figure}

The medium room data samples have a starting MAE of 0.1242 m and a standard deviation of $\sigma=0.1642$ m. On the other hand, the starting MAE of explicitly labeled LoS samples is 0.0594 m.

In Fig. \ref{fig:hist_pred_res} are shown the results obtained by the reference architecture trained with the setting illustrated in subsection \ref{experimental_setting}. It is possible to notice how the network is able to almost completely compensate the offset of the original range error and reduce the standard deviation of $34.1\%$. Moreover, as summarized in Table \ref{tab:optimization_results}, the network can easily detect LoS input signals and apply a small correction factor that takes into account the multipath effect. That is proved by the residual error MAE that has a percentage improvement of $25.1\%$. On the other hand, MAE for NLoS signals is improved by $44.7\%$, reducing the error to 0.0697 m, near the actual precision of DWM1000 boards \cite{jimenez2016comparing}.

In the upper part of Table \ref{tab:optimization_results}, three simple models (SVM, MLP and CNN-1D) are included as a reference. For support vector machine (SVM) and MLP, we adopt the six hand-crafted
features described in \cite{marano2010nlos,mazuelas2018soft}.  We use radial basis function as the kernel for our SVM and a 3-layer architecture with 64 hidden neurons for the MLP. Instead, for \cite{bregar2018improving}, we feed the network with 152 bins, and we set the hyperparameters suggested in the article. It is noticeable how REMnet has better performances than other literature methodologies even with a highly efficient architecture.

Finally, for a matter of completeness, in Table \ref{tab:cross_validation}  are presented results obtained with a cross-validation of the three different room sizes. In accordance with conclusions of Section \ref{dataset_analysis}, REMnet achieves comparable range error mitigation in the three different configurations.

\begin{table}[]
\begin{tabular}{lrrr}
\bottomrule
                        & \multicolumn{1}{c}{SR} & \multicolumn{1}{c}{MR} & \multicolumn{1}{c}{BR} \\ \hline
Training Samples        & 31632                   & 36023                   & 30811                   \\ 
Test Samples {[}NLoS{]} & 17601                   & 13210                   & 18422                   \\ 
Test Samples {[}LoS{]} & 4691                    & 4691                    & 4691                    \\ 
$\sigma_{obs}$ {[}NLoS{]}             & 0,1508                  & 0,1603                  & 0,1851                  \\ 
$\sigma_{res}$ {[}NLoS{]}            & 0,1131                  & 0,1057                  & 0,1204                  \\ 
$\mu_{obs}$  {[}NLoS{]}           & 0,0881                  & 0,1244                  & 0,1057                  \\ 
$\mu_{res}$  {[}NLoS{]}              & 0,0058                  & 0,0153                  & 0,0171                  \\ 
MAE {[}NLoS{]}          & 0,0638                  & 0,0687                  & 0,0702                  \\ 
MAE {[}LoS{]}           & 0,0462                  & 0,0445                  & 0,044                   \\ 
$R^2$ {[}NLoS{]}           & 0,5793                  & 0,5607                  & 0,5                     \\ 
$R^2$ {[}LoS{]}            & 0,4005                  & 0,3483                  & 0,444                   \\ \bottomrule
\end{tabular}
\caption{REMnet cross-validation results with the three different room sizes, small room (SR), medium room(MR) and big room(BR). Each column presents metrics for NLoS and LoS signals for the room escluded by the training procedure.}
\label{tab:cross_validation}
\end{table}

\subsubsection{Power and latency optimization}

Range error mitigation should be performed directly on the platform where the UWB tag is attached. So, energy consumption and computational power play a decisive role in the applicability of the proposed methodology. However, real-time range mitigation with the whole CIR could be very computational intensive \cite{zeng2018nlos}. Consequently, to comply with cost, energy, size, and computational constraints, we investigate the effects of optimization, detailed in Section \ref{methodology}, on the network's accuracy instead of manually extracting a reduced number of features from the CIR.

Graph optimization techniques and different weight quantization levels are both examined, starting from the pre-trained reference network. In Table \ref{tab:optimization_results} are summarized the performances of the model after different optimization processes. Even if there is a degradation of the overall metrics, these changes are mostly negligible. Moreover, it is possible to notice that the full-integer quantization, generally producing a size reduction and speed-up of $75\%$, decreases the NLoS MAE only of the $3\%$ if carried out with awareness training. That opens the possibility to achieve effective range mitigation with an almost negligible impact on the overall application. Indeed, extreme weight quantization implies a smaller model size with less memory usage, an important latency reduction, and the possibility of using highly efficient neural accelerators.

\subsubsection{Inference results}

\begin{figure}
    \centering
    \includegraphics[scale=0.31]{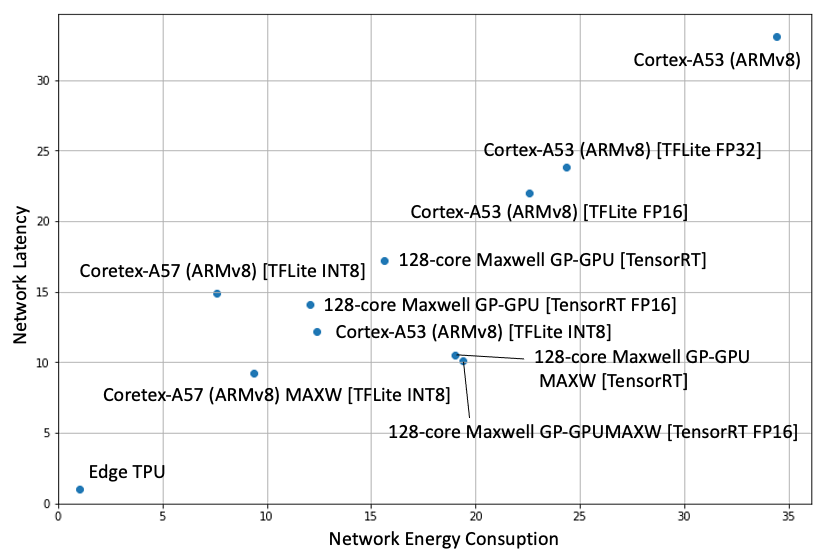}
    \caption{Energy and latency are two important constraints for an effective range error mitigation. Indeed, error correction is performed progressively over all received anchor signals on board of the platform connected with the tag. Without an highly optimized and efficient correction model, range error mitigation would not be applicable.}
    \label{fig:energy_vs_latency}
\end{figure}

In this section, we test different optimized networks on several devices and hardware accelerators, annotating power and computational request. The choice for the selected microchips is made by performing a market evaluation at the time of writing that considers common computational boards for indoor navigation. Indeed, robotic platforms are usually endowed with Linux-capable Arm Cortex-A CPUs or powerful co-processors and accelerators such as Nvidia GP-GPUs, Visual Processing Units (VPUs), or Tensor Processing Units (TPUs). We adopt two standard libraries for network deployment, TensorFlow-Lite\footnote{https://www.tensorflow.org/lite} and TensorRT\footnote{https://developer.nvidia.com/tensorrt} to produce the optimized models. Both are directly integrated into the TensorFlow framework and are specifically conceived to target different hardware platforms. In particular, we target Cortex-A53, A57 processors, and Edge TPU with TF-Lite and the Nvidia RTX 2080 and 128-core Maxwell GP-GPUs with TensorRT.

\begin{table*}[]
\resizebox{\textwidth}{!}{%
\begin{tabular}{llllllllll}
\toprule
Device           & G.O. & W.P. & Latency {[}ms{]}                  & Latency$_{4\;batch}$ {[}ms{]} & $V_{al}$ {[}V{]} & $I_{idl}$ {[}A{]} & $P_{run}$ {[}W{]} & $E_{net}$ {[}mJ{]} & Size {[}KB{]} \\ \hline
RTX 2080         & N    & FP32 & \multicolumn{1}{c|}{19.7 $\pm$ 0.23} & 19.3 $\pm$ 0.24        & N.A.          & N.A.           & 32             & 617.6           & 250.0         \\ 
                 & Y    & FP32 & 0.69 $\pm$ 0.13                      & 0.69 $\pm$ 0.16        & N.A.          & N.A.           & 20             & 138.0           & 613.0         \\ 
                 & Y    & FP16 & 0.54 $\pm$ 0.09                      & 0.51 $\pm$ 0.02        & N.A.          & N.A.           & 18             & 97.2            & 615.0         \\ 
Cortex-A53       & N    & FP32 & 16.9 $\pm$ 0.03                      & 17.2 $\pm$ 0.05        & 5.0           & 0.4            & 1.0            & 17.2            & 250.0         \\ 
                 & Y    & FP32 & 12.2 $\pm$ 0.03                      & N.A.                & 5.0           & 0.4            & 1.0            & 12.2            & 40.7          \\ 
                 & Y    & FP16 & 11.2 $\pm$ 0.03                      & N.A.                & 5.0           & 0.4            & 1.0            & 11.2            & 33.9          \\ 
                 & Y    & INT8 & 6.23 $\pm$ 0.02                      & N.A.                & 5.0           & 0.4            & 1.0            & 6.2             & 32.7          \\ 
Cortex-A57       & Y    & INT8 & 7.63                              & N.A.                & 5.0           & 0.5            & 0.8            & 3.81            & 32.7          \\ 
                 & Y    & INT8 & 4.71                              & N.A.                & 5.0           & 0.5            & 1.0            & 4.7             & 32.7          \\ 
128-core Maxwell & Y    & FP32 & 8.78 $\pm$ 0.09                      & 9.03 $\pm$ 0.1         & 5.0           & 0.67           & 0.9            & 7.8             & 615.0         \\ 
                 & Y    & FP16 & 7.22 $\pm$ 0.08                      & 7.43 $\pm$ 0.05        & 5.0           & 0.67           & 0.9            & 6.04            & 613.0         \\ 
                 & Y    & FP32 & 5.36 $\pm$ 0.05                      & 5.29 $\pm$ 0.05        & 5.0           & 0.5            & 1.8            & 9.7             & 615.0         \\ 
                 & Y    & FP16 & 5.18 $\pm$ 0.04                       & 5.39 $\pm$ 0.05        & 5.0           & 0.5            & 1.0            & 4.7             & 613.0         \\ 
Edge TPU         & Y    & INT8 & 0.51 $\pm$ 0.1                       & N.A.                & 5.0           & 0.59           & 0.7            & 0.5             & 70.54         \\ \bottomrule
\end{tabular}}
\caption{Comparison between different devices energy consumption and inference performances. Graph optimization (G.O.) and weight precision (W.P.) reduction further increase the capability of our already efficient neural network design helping to deal with energy, speed, size and cost constraints.}
\label{tab:inference_resutls}
\end{table*}

Experimentation results are summarized in Table \ref{tab:inference_resutls}. It is possible to notice that, due to the high efficiency of the proposed architecture, all configurations satisfy a sufficient inference speed compliant for an effective range error mitigation solution. Nevertheless, the different optimization techniques applied have a high impact on the energy consumed by the network. Indeed, considering experimentations performed with the Cortex-A53, optimization can reduce the energy consumption by nearly a factor of three, starting with an initial value of 17.2 mJ to barely 6.2 mJ with a reduction of $64\%$. Moreover, the model size is greatly reduced from 250 KB to 32.7 KB. That implies a smaller storage size and less RAM at run-time, freeing up memory for the main application where UWB localization is needed. Finally, as further highlighted by Fig. \ref{fig:energy_vs_latency} and the results of the previous subsection, the Edge TPU neural accelerator with a full-integer quantized aware training model is the preferable solution for deployment. With only 0.51 ms of latency and 0.5 mJ energy consumption, it barely impacts the performance of the overall application, allowing to exploit duty cycling and energy-saving techniques. Indeed, as stated by our proposed methodology section, the already efficient design of our architecture, in conjunction with 8-bit weight precision and graph optimization techniques, makes deep learning a feasible solution for an effective range error mitigation for UWB at the edge.


\subsubsection{Trilateration results}
\begin{figure}
    \centering
    \includegraphics[scale=0.44]{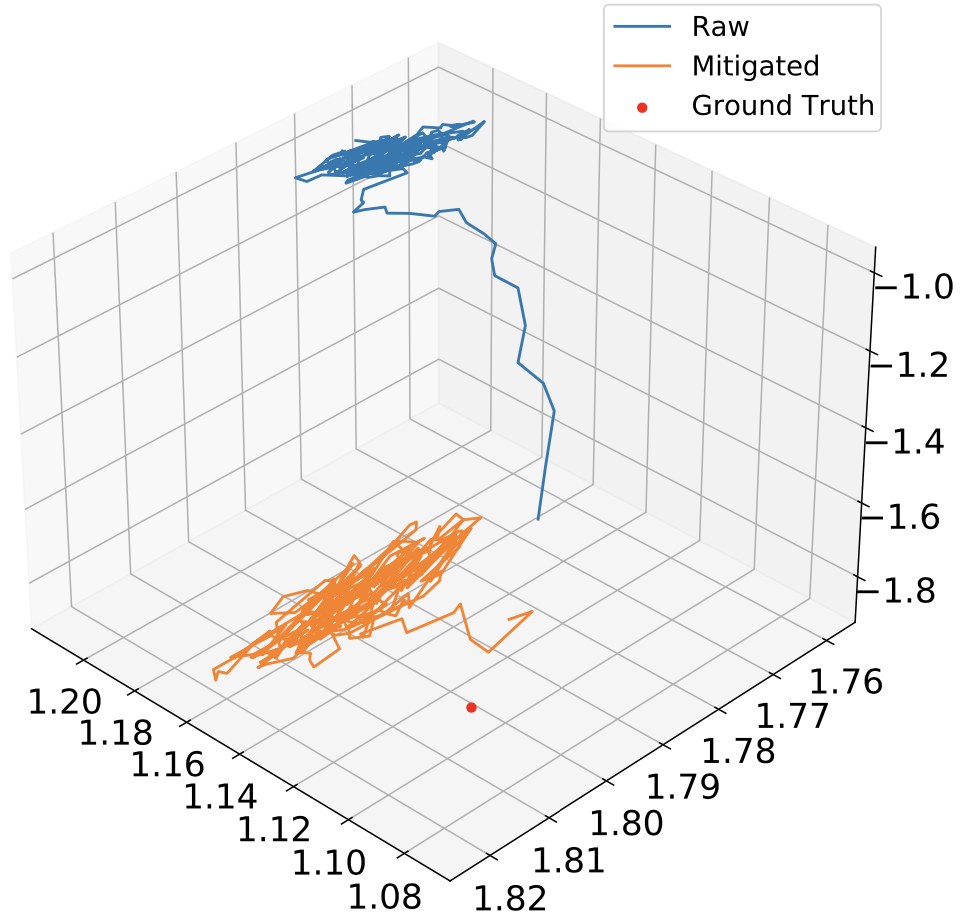}
    \caption{Comparison between position estimations of a fixed tag in NLoS conditions. In light blue the results obtained from raw range measurements, in orange the ones achieved with our quantization aware mitigation model.}
    \label{fig:pos}
\end{figure}
As described in \ref{experimental_setting}, the effect of the proposed method is lastly verified by using the full-integer quantization aware model for a 3D positioning task, in which the results obtained from raw range estimates are compared to the ones achieved with our mitigation model in the loop. The results are summarized in Table \ref{tab:pos}, while Fig. \ref{fig:pos} gives a graphical representation of the NLoS results. Regarding the LoS case, the positioning system already achieves a good precision by itself with very low range MAE and, consequently, a low position MAE. In this case, the effect of mitigation is irrelevant, causing a slight increment of ranging error but a slight decrease in positioning error. So, as expected, the model learns to apply very slight corrections to LoS samples, avoiding worsening already good measurements. Instead, the NLoS scenario shows a significant improvement, as the range MAE is more than halved, reaching a value that is comparable to the LoS case and confirming the results shown in \ref{quantitative_results}.
Consequently, the error on the position estimation is strongly reduced, going from 57.7 cm to 18.2 cm. Although the final accuracy is still significantly higher than the one found in the LoS case, a reduction of $68\%$ is considered a significant result. Indeed, our approach allows achieving a suitable precision for many kinds of indoor robotic applications showing good generalization to unknown environments.

\begin{table*}[]
\centering
\begin{tabular}{@{}lllll@{}}
\toprule
                                & \multicolumn{2}{c}{LOS}                  & \multicolumn{2}{c}{NLOS}                 \\ \midrule
                                & Range MAE {[}m{]} & Position MAE {[}m{]} & Range MAE {[}m{]} & Position MAE {[}m{]} \\
Raw UWB Measurements            & 0.0388            & 0.0703               & 0.1129            & 0.5772               \\
Full-integer Aware Quantization & 0.0465            & 0.0679               & 0.0571            & 0.1817               \\ \bottomrule
\end{tabular}
\caption{Results obtained from the positioning test in the medium room, that is not used for the training of the model. For each test, the mean absolute error is reported for both the range estimates and the final position result, in order to highlight the effect of the former on the latter.}
\label{tab:pos}
\end{table*}

\section{Conclusions}
We introduced REMnet, a novel representation learning model accurately designed to constitute an effective range error mitigation solution. Moreover, we proposed a set of optimization techniques to enhance its efficiency and computational results further. Extensive experimentation proved the effectiveness of our methodology and generality over disparate scenarios.
Further works will aim at integrating the deep learning architecture on an ultra-low-power microcontroller directly placed on the UWB device.

\section*{Acknowledgements}
This work has been developed with the contribution of the Politecnico di Torino Interdepartmental Centre for Service Robotics PIC4SeR\footnote{https://pic4ser.polito.it} and SmartData@Polito\footnote{https://smartdata.polito.it}. Moreover, it is partially supported by the Italian government via the NG-UWB project (MIUR PRIN 2017).


\bibliographystyle{IEEEtran}
\bibliography{mainArXiv}

\vfill\break

\clearpage

\begin{IEEEbiography}[{\includegraphics[width=1in,height=1.25in,clip,keepaspectratio]{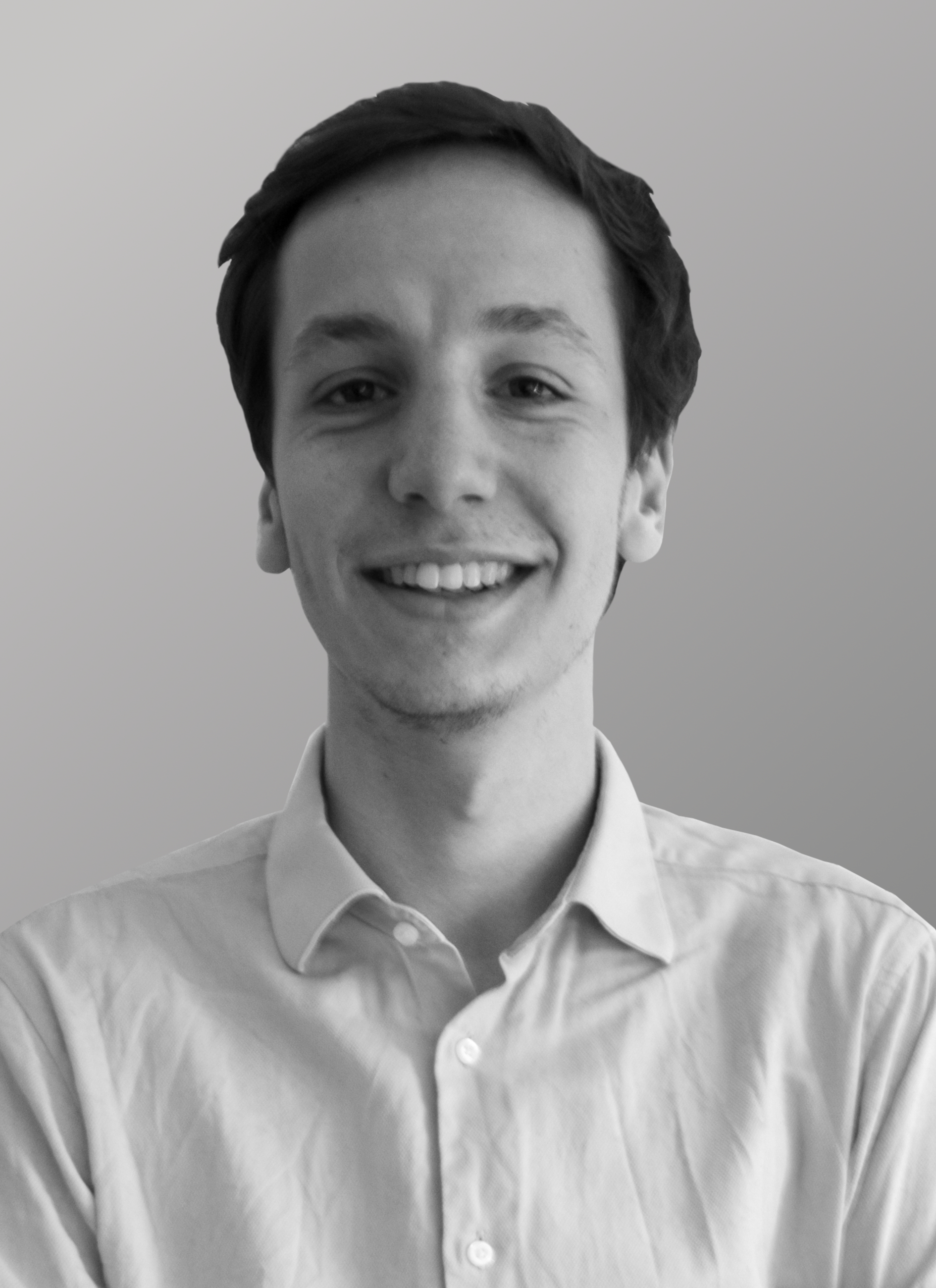}}]{Simone Angarano} is a research fellow at PIC4SeR (\url{https://pic4ser.polito.it/}). He achieved a Bachelor's Degree in Electronic Engineering in 2018 and a Master's Degree in Mechatronic Engineering in 2020 at Politecnico di Torino, presenting the thesis "Deep Learning Methodologies for UWB Ranging Error Compensation". His research focuses on Machine Learning for robotic applications in everyday-life contexts.
\end{IEEEbiography}

\begin{IEEEbiography}[{\includegraphics[width=1in,height=1.25in,clip,keepaspectratio]{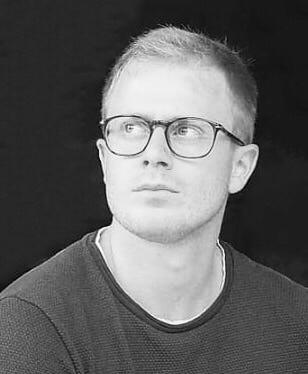}}]{Vittorio Mazzia} is a Ph.D. student in Electrical, Electronics and Communications Engineering working with the two Interdepartmental Centres PIC4SeR (\url{https://pic4ser.polito.it/}) and SmartData (\url{https://smartdata.polito.it/}). He received a master's degree in Mechatronics Engineering from the Politecnico di Torino, presenting a thesis entitled "Use of deep learning for automatic low-cost detection of cracks in tunnels," developed in collaboration with the California State University. His current research interests involve deep learning applied to different tasks of computer vision, autonomous navigation for service robotics, and reinforcement learning. Moreover, using neural compute devices (like Jetson Xavier, Jetson Nano, Movidius Neural Stick) for hardware acceleration, he is currently working on machine learning algorithms and their embedded implementation for AI at the edge. 
\end{IEEEbiography}

\begin{IEEEbiography}[{\includegraphics[width=1in,height=1.25in,clip,keepaspectratio]{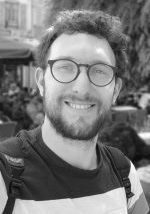}}]{Francesco Salvetti} is currently a Ph.D. student in Electrical, Electronics and Communications Engineering in collaboration with the two interdepartmental centers PIC4SeR (\url{https://pic4ser.polito.it/}) and Smart Data (\url{https://smartdata.polito.it/}) at Politecnico di Torino, Italy. He received his Bachelor's Degree in Electronic Engineering§ in 2017 and his Master's Degree in Mechatronics Engineering in 2019 at Politecnico di Torino. He is currently working on Machine Learning applied to Computer Vision and Image Processing in robotics applications.
\end{IEEEbiography}

\begin{IEEEbiography}[{\includegraphics[width=1in,height=1.25in,clip,keepaspectratio]{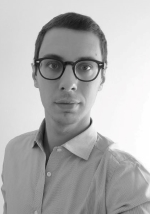}}]{Giovanni Fantin} is a research fellow at PIC4SeR (\url{https://pic4ser.polito.it/}). In 2019, he achieved the Master's Degree in Mechatronics Engineering at Politecnico di Torino discussing the thesis "UWB localization system for partially GPS denied robotic applications". He is currently working on a PRIN (progetto di rilevante interesse nazionale) about new generation ultra-wideband technologies with a particular focus on multi-robot cooperation to perform localization.
\end{IEEEbiography}

\begin{IEEEbiography}[{\includegraphics[width=1in,height=1.25in,clip,keepaspectratio]{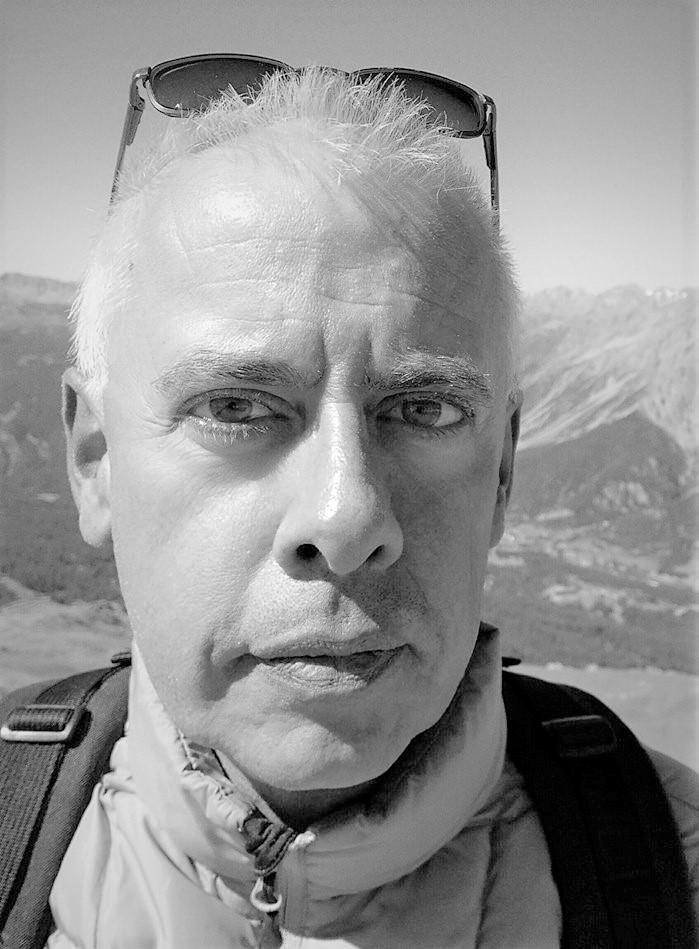}}]{Marcello Chiaberge} is currently an Associate Professor within the Department of Electronics and Telecommunications, Politecnico di Torino, Turin, Italy. He is also the Co-Director of the Mechatronics Lab, Politecnico di Torino
(\url{www.lim.polito.it}), Turin, and the Director and the Principal Investigator of the new Centre for Service Robotics (PIC4SeR, \url{https://pic4ser.polito.it/}), Turin. He has authored more than 100 articles accepted in international conferences and journals,
and he is the co-author of nine international patents. His research interests include
hardware implementation of neural networks and fuzzy systems and the design and implementation of reconfigurable real-time computing architectures.  \end{IEEEbiography}
\vfill

\end{document}